# Data-driven method for real-time prediction and uncertainty quantification of fatigue failure under stochastic loading using artificial neural networks and Gaussian process regression

Maor Farid

*Massachusetts Institute of Technology, 77 Massachusetts Ave., Cambridge, MA 02139, United States*
*Technion – Israel Institute of Technology, Haifa 3200003, Israel*
*faridm@mit.edu*

## Abstract

Various engineering systems such as naval and aerial vehicles, offshore structures, and mechanical components of motorized systems, are exposed to fatigue failures due to stochastic loadings. Methods for early failure prediction are essential for engineering, military, and civil applications. In addition to the prediction of time to failure (TtF), uncertainty quantification (UQ) is of major importance for real-time decision-making purposes. Usually, time domain or frequency domain methods are used for fatigue prediction, such as rainflow counting and Miner's rule or Dirlik's method. However, those methods suffer from over-simplistic modeling and inaccurate failure predictions under stochastic loadings. During the last years, several data-driven models were suggested for offline fatigue failure. However, most of them are not capable of both accurate real-time fatigue prediction and UQ. In the current work, a probabilistic data-driven model is introduced. A hybrid architecture of a fully-connected artificial neural network (FC-ANN) and Gaussian process regression (GPR) is proposed to ensure enhanced predictive abilities and simultaneous UQ of the predicted TtF. The real-time prediction and UQ performances of the suggested model are validated using both synthetic and experimental data. This novel hybrid method is fully data-driven and extends the forecasting capabilities of existing time-domain and machine learning-based methods for fatigue prediction. It paves the way towards the development of a preventive system that provides real-time safety and operational instructions and insights for structural health monitoring (SHM) purposes, allowing prevention of environmental damage, and loss of human lives.

*Keywords:* Real-time fatigue prognosis, data-driven methods, machine learning, artificial neural networks, Gaussian process regression, Bayesian inference, uncertainty quantification.

## 1. Introduction

Fatigue failure refers to the malfunctioning of a mechanical component due to the weakening of its material under oscillatory loading below the ultimate tensile strength of the component. Fatigue is considered as one of the main reasons for mechanical failures in machine components, aerospace systems, and offshore structures [1]. The failure results in localized and progressive damage and in crack growth. Fatigue damage is cumulative over time, and even though it can be assessed by non-destructive tests [2], it still can take place at unexpected timing, leading to hazardous consequences. Notable examples include the capsizing of Alexander L. Kielland semi-submersible drilling rig in March 1980 [3, 4], and the crash of the El-Al Flight 1862 in October 1992 [5] causing the death of dozens of workers, passengers, and crew members. Hence, reliable estimation of the current state of the cumulative damage and prediction of the time to failure (TtF) are essential for SHM and failure prognosis purposes [6, 7].

Fatigue damage under an oscillatory loading is usually estimated by either time-domain or frequency-domain approaches. The most widely-used time-domain approach is based on the rainflow counting method for decomposing the stochastic signal to its underlying amplitudes and corresponding number of cycles, followed by applying the linear cumulative Miner's rule, also known as the Palmgren-Miner linear damage hypothesis [8, 9] for estimating the resulting cumulative damage, as shown in Eq. (1). Frequency domain methods, such as those introduced by Dirlik [10] and Petrucci and Zuccarello [11], use probability density functions with parameters that are tuned with respect to the rainflow counting and the Miner's rule [12, 13, 14]. Due to multiple observatory studies, rainflow counting and Miner's rule are considered more reliable and accurate in comparison to the frequency domain methods. However, Miner's rule is considered as an over-simplistic failure criterion that overlooks failure mechanisms at the material level [13]. Moreover, its accuracy deteriorates drastically as the component is exposed to broader-banded stochastic excitations, leading to substantial disagreements between theoretical and experimental results reported in the literature [15].

$$D(t) = \sum_{i=1}^{N_k} \frac{n_i(t)}{N_{f,i}}, \quad N_{f,i} = \left(\frac{S_{a,i}}{A\bar{\alpha}}\right)^{\frac{1}{b}}, \quad \bar{\alpha} = 1 - \frac{x_m}{\sigma_{uts}} \quad (1)$$

Here $D(t)$ is the estimated cumulative damage fraction. According to Miner's rule, failure occurs when $D(\tau) = 1$, where $\tau$ is the estimated failure time (FT). Variable $N_k$ is the number of amplitudes considered in the rainflow counting method. Variable $S_{a,i}$ is the stress amplitude that leads to failure after $N_{f,i}$ loading cycles, and $n_i(t)$ is the number of loading cycles of amplitude $S_{a,i}$ counted by time $t$. Parameters $A, b$ and $\sigma_{uts}$ are the fatigue strength, fatigue exponent and ultimate tensile stress of the component material, respectively. Variable $x_m$ is the mean stress of the loading signal.

Fatigue life is defined as the number of stress cycles that a specimen sustains before failure. In the current work, we focus on TtF prediction rather than fatigue life as required for SHM purposes and real-time decision making in engineering systems. One of the main challenges in the prediction of fatigue failures is the uncertainty in the material parameters at the microscopic level that leads to an inherent noise and uncertainty in the data and inconsistencies in the resulting failure times (FTs). Hence, due to the uncertainty associated with the material and mechanical properties of the component, systematic experimental noise, and statistical properties of the stochastic loading, the TtF should be treated as a probabilities quantity, which is characterized by a nominal predicted value and a corresponding level of uncertainty. Uncertainty quantification (UQ) becomes critical when it comes to real-time decision-making in systems for which sudden changes in the operating regime might lead to hazardous consequences. For example, a suggestion for emergency landing of a fighter jet due to a predicted failure in absence of a near airport. In this case, the certainty level is crucial for choosing between continuation or cessation of the mission.

In the absence of closed-form expressions that link the system's mechanical parameters and signal properties with the TtF, and due to the abundance and stochastic nature of experimental and numerical data, data-driven methods are good candidate models for fatigue prognosis. Machine learning (ML) methods are a set of numerical models that fit a function to a given dataset by tuning a large set of model parameters in a process called learning or training. During the last decade, learning algorithms have gained popularity due to their versatility, the substantial availability of rich datasets, and high-end computational resources. Those models outperformed conservative methods in various tasks, such as image classification and speech recognition, due to their well-known interpolation and generalization capabilities. ML algorithms were proven to have a good ability to capture underlying patterns and correlations in measured data even when the underlying physical rules governing the system's behavior are obscure or unknown. Examples include quantitative analysis of ceramic's fracture surfaces using convolutional neural networks [16] and identification of underlying equations of motion of dynamical systems from measured data [17, 18]. Therefore, applying machine learning approaches for fatigue prognosis while utilizing the availability of high-end computational abilities and rich datasets obtained from experiments [19], numerical simulations [20, 21], or analytical analysis, is a promising strategy for cumulative damage estimation. However, limited amount of studies focused on ML-based real-time fatigue prediction under stochastic loading [22, 23, 24, 25, 26, 27]. Most of the suggested methods utilized rather simple ML-based predictive models which are applicable only in an offline fashion, i.e. in a retrospective reading of the full loading signals, and thus are not applicable for real-time purposes. Some of them are applicable only for zero-mean stochastic loadings. Moreover, those models aim to predict the fatigue life of the component or the cumulative fatigue damage and do not provide a quantitative measure of the level of uncertainty associates with the predicted value.

The main goal of the current work, is to develop a fully data-driven method for real-time fatigue failure prediction and UQ under non-zero-mean stochastic loading of unknown underlying power spectral density (PSD). In particular, we introduce a hybrid learning model that combines both a fully-connected artificial neural network (FC-ANN) and Gaussian process regression (GPR) [28, 29, 30, 31, 32] for enhanced FT prediction and built-in UQ capabilities. FC-ANNs have a well-known ability to identify latent relations between a set of inputs and a quantity of interest, in this case, system and forcing properties and the predicted TtF. The main advantage of employing GPR is the simultaneous estimation of the output value and the associated error/uncertainty. By using GPR, the formulated hybrid model can generate a probability density function (PDF) over the range FTs that embodies both the predicted TtF and the corresponding uncertainty associated with various sources. We also purpose a Bayesian inference-based scheme that utilizes real-time measured data for constantly updating the posterior knowledge of the predictive model. Based on the resulting posterior PDF, both the predicted FT and a confidence interval (CI) are obtained. The size of the CI is controlled by desired confidence level (CL), that can be modified on user demand and in accordance to the sensitivity of the system. This novel approach is, therefore, suitable for real-time decision-making in uncertain conditions. Moreover, it is applicable for data obtained from various sources, such as experiments, numerical simulations, and approximate analytical methods. Finally, we evaluate the performances of the proposed method for real-time fatigue prognosis for both synthetic and experimental unseen measured stochastic signals.

This paper is structured as follows. In Section 2 we discuss the input features for the learning model. In Section 3 we introduce the hybrid probabilistic predictive model. In the same section, we give an overview of fully-connected artificial neural networks and Gaussian process regression and describe the contribution of each of those modules to the overall performances of the hybrid model. The proposed methodology is demonstrated in Section 4 where it is applied on an unseen stochastic augmented data in real-time. In Section 5 we validate the hybrid model on experimental data obtained from mechanical systems that experienced fatigue failure under stochastic loadings. Finally, Section 6 provides a summary and brief discussion of the results obtained, possible engineering applications, and future research directions.

## 2. Input features for the learning model

Before a learning model can be designed or applied, the most informative input variables have to be chosen in advance. Those variables, also called features, have to represent the relevant physical properties of the mechanical component and statistical properties of the stochastic signal. The set of all features that correspond to a given mechanical component in the dataset and that was measured in a given time $t$ is called the temporal features vector $\mathbf{u}(t)$ and serves as the input vector of the predictive model. The measured signal can be any physical measure from which the stress at the critical section of the forced component can be inferred, such as displacement, acceleration, or stress at the fatigue-prone location or in its vicinity. As mentioned in the previous section, any data from various sources can be used to train the model and test its performances, such as data obtained from experimental, numerical, and/or approximate analytical models, as long as data of the same type is used both for training the model and evaluating its results. Generally, if the component of interest is subjected to stochastic excitations of various characteristics during its lifetime, it is recommended to include signals that were generated from a set of relevant PSDs to enhance the generalization ability of the learning model. Otherwise, a single PSD can be considered to avoid the necessity of a large dataset and extensive computation time. In the current work, without loss of generality, we use stress signal at the critical section of the mechanical component for reasons of convenience and explainability of the results. The features are generated from a uniform probability distribution over the parameter ranges shown in Table 1. The stochastic signals are then generated using Eq. (2)-(3), where the amplitudes of the Fourier series are generated randomly from Gaussian-like PSD. The resulting stochastic signals are assumed to be ergodic and statistically stationary. Both the normalization term in the denominator and scaling factor $k_s$ in Eq. (2) are used to prevent ultra low-cycle fatigue due to stresses that exceed the ultimate tensile stress of the material $\sigma_{uts}$. The scaling parameter is generated from a uniform distribution over the range shown in Table 1.

$$\begin{aligned} x(t) &= x_m + (k_s\sigma_{uts} - |x_m|)\chi(t) \\ \chi(t) &= \frac{\sum_{i=1}^{N_f} \sqrt{2G(f_i)\Delta f} \cos(2\pi f_i t + \phi_i)}{\max\left(\sum_{i=1}^{N_f} \sqrt{2G(f_i)\Delta f} \cos(2\pi f_i t + \phi_i)\right)} \end{aligned} \quad (2)$$

Here $\chi(t)$ is a normalized stochastic signal which is bounded between $\pm 1$ and generated from an underlying PSD $G(f)$. Variables $x_m$ is the mean of signal $x(t)$ and $k_s$ is a scaling factor. Variables $f_i$ and $\phi_i$ are the constructing frequencies and phases of stochastic signal $\chi(t)$. All variables in Eq. (2) are randomly generated from a uniform distribution over the ranges shown in Table 1. Thus, $x(t)$ can be treated as a realizations of a random variable, $x(t) \sim X(t,\zeta)$.

$$G(f) = \frac{1}{\sqrt{2\pi}\sigma_G} \exp\left(\frac{-(f-\mu_G)^2}{2\sigma_G^2}\right) \quad (3)$$

As one can see in (2), the PSD $G(f)$ does not include a magnitude factor. This is because the intensity of the stochastic signal $x(t)$ which is usually dictated by the magnitude of the PSD, is now fully-defined by the scaling factor $k_s$. The range of scaling factor $k_s$ was chosen to ensure that the generated signal will not exceed 85% of the ultimate tensile stress of the material $\sigma_{uts}$, and lead to low-cycle fatigue (LCF) [33] or ultra low-cycle fatigue (ULCF) [34]. The features vector includes the material coefficient that characterizes the material resistance to fatigue, i.e. the fatigue strength $A$, fatigue exponent $b$, and ultimate tensile strength $\sigma_{uts}$ [13]. Next, it is well-known that the mean of the measured signal has a significant effect on the TtF. Since the mean of the signal is assumed to be unknown we estimate it with time using Eq. (4):

$$\bar{x}(t) \equiv \mathbb{E}\left[x(t)\right] = \frac{1}{t}\int_0^t x(t')dt' \tag{4}$$

Here $\mathbb{E}$ denotes expectation. The spectral information in the measured stochastic signal $x(t)$ is embodied in its underlying power spectral density (PSD) Eq. (5) describes the variation of power content of a signal versus frequency.

$$G(f|t) = \mathbb{E}\left[|X(f|t)|^2\right], \quad X(f|t) = \int_0^t x(t)e^{2\pi ft}dt \tag{5}$$

Here $X(f|t)$ is the Fourier transform of signal $x(t)$ for a given measurement time $t$. Since we assume that the PSD of the signal is unknown a-priori, it is estimated in real-time using Welch's method, which computes the periodograms for overlapping segments of the given signal and averages over the results, allowing more accurate PSD estimation with time $G(f|t)$ for longer measurement durations [35]. The spectral information in the PSD can be reduced to a handful of representative quantities, called spectral moments, Eq. (6).

$$m_i(t) = \int_0^\infty f^i G(f|t)df, i = 0, 1, 2, 4 \tag{6}$$

The four moments shown in Eq. (6), embody most of the spectral information in the PSD and especially the bandwidth of the stochastic loading, as one can learn from the expression of the irregularity factor $\bar{\gamma}$ of a PSD expresses as Eq. (7). For broad-banded and narrow-banded loadings, it tends to zero and one, respectively. Any higher odd or even moments will not be adding new information regarding the frequency content of the signal. Since the PSD $G(f|t)$ is evaluated with time in a fixed rate of $f_s$, the spectral moments are time-dependent as well $m_i(t)$.

$$\bar{\gamma}(t) = \frac{m_2(t)}{\sqrt{m_0(t)m_4(t)}} \tag{7}$$

The bandwidth of a stochastic signal has a direct effect on the distribution of the signal local extrema around the mean value of the signal. Since fatigue failure is dictated by the amplitudes of the loading cycles, this distribution has a significant effect on the cumulative damage and TtF. This distribution is given by the following expression:

$$\rho(\xi) \equiv P(|M| \leq \xi) = \int_{-\xi}^{\xi} f(\bar{\xi})d\bar{\xi} \tag{8}$$

Here $f$ and $P$ are PDF and probability, respectively. Random variable $M$ is drawn from a set $\mathcal{M}$ of all local extrema in the signal $x(t) - \bar{x}(t)$.

random variable $M$ is a drawn from the set of all local extrema $\mathcal{M}$ of signal $x(t) - \bar{x}(t)$, i.e. $M \sim \mathcal{M}$. Variable $\xi$ is an arbitrary amplitude value and $f(\xi)$ is the PDF over probable $\xi$ values. Function $\rho(\xi)$ describes the probability that a local maximum $M$ is bounded between $\pm \xi$. Assuming ergodic and statistically stationary stochastic signal, for broad-banded and narrow-banded PSDs, the PDF $f(\xi)$ correspond to the Rayleigh distribution and the multi-modal

Gaussian distribution, respectively. Comparison between theoretical and numerical results are shown in Fig. 1.

$$f_{NB}(\xi) = \frac{\xi}{\tilde{\sigma}^2}\left(\exp\left(\frac{-\xi^2}{2\tilde{\sigma}^2}\right)H(\xi) - \exp\left(\frac{-\xi^2}{2\tilde{\sigma}^2}\right)H(-\xi)\right)$$
$$f_{BB}(\xi) = \frac{1}{\sqrt{8\pi}\tilde{\sigma}}\left(\exp\left(\frac{-(\xi-\tilde{\mu})^2}{2\tilde{\sigma}^2}\right) + \exp\left(\frac{-(\xi+\tilde{\mu})^2}{2\tilde{\sigma}^2}\right)\right) \quad (9)$$
$$\tilde{\mu} = \mathbb{E}\left[\mathcal{Q}\{x(t)-\bar{x}\}\right], \tilde{\sigma} = \sqrt{\mathbb{E}\left[(\mathcal{Q}\{x(t)-\bar{x}\}-\tilde{\mu})^2\right]}$$

Here $H(\xi)$ is the Heaviside step function, and $\mathcal{Q}$ is an operator that maps a continuous signal to a set of its local maxima. Parameters $\tilde{\mu}$ and $\tilde{\sigma}$ are the mean and standard deviation of the distribution of the local maxima.

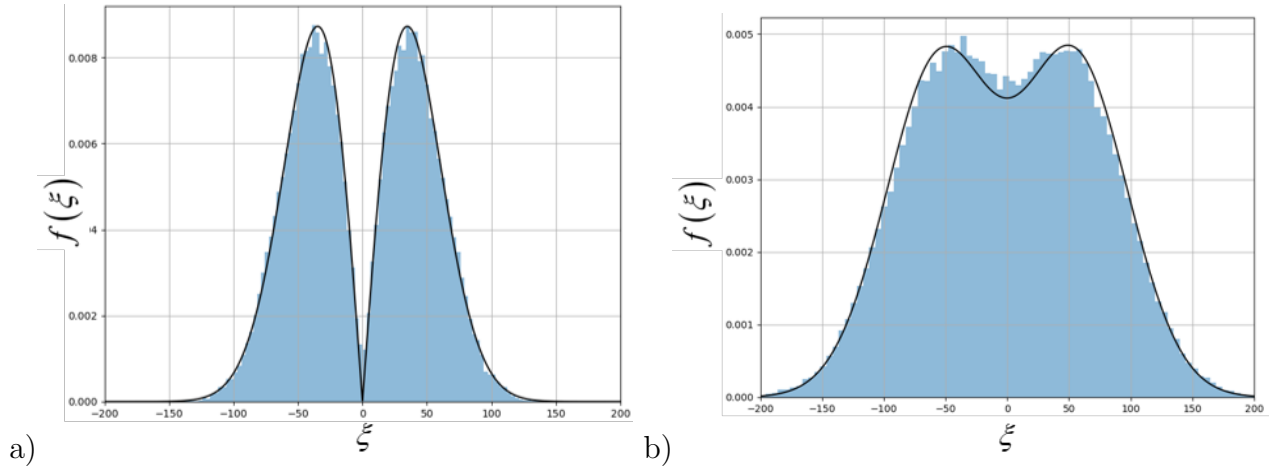

Figure 1: Comparison between histogram and theoretical PDFs of amplitudes/local maxima distribution of stochastic signal around its mean, i.e. $x(t) - \bar{x}, x(t) \sim X(t, \zeta)$; a) narrow-banded signal obtained for Gaussian-like PSD in Eq. (2) for $\mu_G = 500$Hz, $\sigma_G = 20$Hz, b) uniform PSD. Both signals consist of $N_f = 20$ frequencies in the range $f \sim \mathcal{U}[0, 1000]$Hz.

For the case of narrow-based PSD, the following relation between mean and STD holds $\tilde{\sigma}\sqrt{2/\pi}\tilde{\mu}$. Then, according to Eq. (8)-(9) we get the corresponding PDFs $\rho(\xi)$:

$$\rho_{NB}(\xi) = 1 - \exp\left(\frac{-\xi^2}{2\tilde{\sigma}^2}\right)$$
$$\rho_{BB}(\xi) = \Phi\left(\frac{\xi+\tilde{\mu}}{\tilde{\sigma}}\right) + \Phi\left(\frac{\xi-\tilde{\mu}}{\tilde{\sigma}}\right) - 1 \quad (10)$$

Here $\Phi(a) = \frac{1}{2}\left[1 + \text{erf}(a/\sqrt{2})\right]$ is the cumulative distribution function of the standard normal distribution. The inverse relations of Eq. (10) describe the $\rho$ percentile amplitude, i.e. the amplitude $\xi$ that is larger than $\rho$-percent of all cycle amplitudes/local maxima of the shifted signal $x(t) - \bar{x}$. This relation can be obtained explicitly only for the narrow-banded case: $\xi_{NB}(\rho) = \tilde{\sigma}\sqrt{\ln[(1-\rho)^{-2}]}$. For the broad-banded case, this relation is obtained numerically $\xi_{BB}(\rho)$.

The value of $\rho$ is tuned by preliminary analysis to optimize the performances of the predictive on a subset of the dataset, as shown in Appendix A. For a given value of $\rho$, the percentile amplitude $\xi$ can be estimated using Eq. (10) and Fig. 2, or estimated directly from the measured signal in real-time, using Eq. (11) which is a discrete form of Eq. (8). The inverse relation $\xi(t|\rho)$ is obtained numerically.

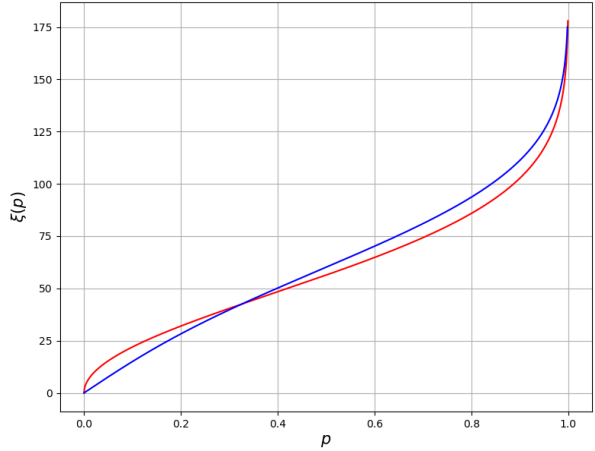

Figure 2: The $\rho$-percentile loading amplitude $\xi$ vs. percentage $\rho$ for both narrow-banded PSD (red) and broad-banded PSD (blue).

$$\rho(\xi) = \frac{|\mathcal{M} \cap [-\xi, \xi]|}{|\mathcal{M}|} \qquad (11)$$

Finally, we obtain a vector that contains indicative information about the properties of the mechanical components and statistical properties of the stochastic signal:

$$\mathbf{u}(t|\rho) = \{A, b, \sigma_{uts}, \bar{x}(t), \xi(t|\rho), m_0(t), m_1(t), m_2(t), m_4(t)\} \qquad (12)$$

The last six features are evaluated in real time using averaging Eq. (4), (6), (11). While the four moments $m_i(t)$ are very distinctive for signals from multiple underlying PSDs, they become redundant when the signals in the dataset belong to a single underlying PSD. In this case, the most indicative features are the mean amplitude $\bar{x}(t)$ and the amplitudes distribution around the mean, i.e. the percentile amplitude $\xi(t|\rho)$. The former cannot be assessed from the PSD, and the latter is a nonlinear combination of the spectral moments of the underlying PSD. Therefore, and in order to reduce the required size of the dataset and enhance computational efficiency and accuracy, we can take the following reduced input vector with five features instead of nine:

$$\mathbf{u}(t|\rho) = \{A, b, \sigma_{uts}, \bar{x}(t), \xi(t|\rho)\} \qquad (13)$$

This case is realistic and typical for mechanical systems that experience one excitation regime that is associated with fatigue failure, for example, components of a machine that is powered by a synchronous electric motor. For simplicity and explainability, in the current work, we will focus on the case of a single PSD and therefore use the reduced features vector. In the current work, preliminary investigation shows a strong correlation between the TtF and the characteristic amplitude that corresponds to the $90^{th}$ percentile, i.e. $\xi(\rho = 0.9) \equiv \xi_{90}$. For convenience, it will be denoted as $\xi$. We generate a dataset that consists of $N = 1500$ statistically stationary and ergodic stochastic signals. All signals were generated from a Gaussian-like PSD using Eq. (2)-(3) for $\mu_G = 150$Hz, $\sigma_G = 500$Hz, and $N_f = 20$ random frequencies and phases generated from a single uniform distribution over the ranges shown in Table 1. Thus, the reduced feature vector is considered. The three material/mechanical coefficient and the actual mean of the signal $x_m$ were randomly generated from a uniform distribution over the ranges shown in Table 1. The $90^{th}$-percentile amplitude $\xi$ and mean amplitude $\bar{x}$ are evaluated from the resulting signal using Eq. (4), (11). The FT is calculated using the rainflow cycle counting method and Miner's rule in Eq. (1). To simulate a systematic random error, a zero-mean Gaussian noise is added to every feature in the input vector $\mathbf{u}(t)$. The magnitude of the additive noise is chosen as $2.5\%$ of the feature range.

| Material/loading parameter | Range |
|---|---|
| Fatigue Strength Coefficient, A, MPa | $1200 - 1500$ |
| Fatigue Strength Exponent, b | $-0.2 - -0.15$ |
| Ultimate Tensile Strength, $\sigma_{uts}$, MPa | $500 - 1000$ |
| True signal mean, $x_m$, MPa | $0 - 250$ |
| Scaling factor, $k_s$ | $0.05 - 0.85$ |
| Constructing frequencies, $f_i$, Hz | $0 - 1000$ |
| Constructing phases, $\phi_i$ | $0 - 2\pi$ |

Table 1: The ranges of material and loading parameters used for generating the dataset.

## 3. Hybrid probabilistic predictive model

In this section, we introduce a probabilistic model for prediction and UQ of fatigue FTs. This hybrid model combines a FC-ANN and a GPR for both enhanced prediction accuracy and simultaneous UQ. The dataset is divided into three subsets: training-set, cross-validation (CV) set, and a test-set, with a ratio of $60:20:20$, respectively. The first is used for tuning the FC-ANN model parameters $\boldsymbol{\theta}$, the second is used for testing the FC-ANN prediction performances, and to train the GPR, and the third for evaluating the real-time prediction and UQ performance of the hybrid predictive model.

### 3.1. Fully-connected artificial neural network

Various learning algorithms have demonstrated good regression and generalization capabilities on datasets of a variety of types. Those mathematical models differ from each other in their architecture, principles of operation, and types of data for which they are applicable. One of the most widely-used and versatile learning models is the fully-connected artificial neural network. The FC-ANN consists of multiple interconnected variables called neurons which are organized in layers, from which stems its name. In virtue of the universal approximation theorem, any continuous (and not necessarily smooth) function can be approximated using a sufficiently complex or deep FC-ANN, where its complexity or depth is determined by the number of neurons and layers of the network. The input vector is fed into the first layer of the network, called the input layer. Then it propagates forward through the network until the output is obtained in the last layer, called the output layer. The size, i.e. number of neurons, in the input and output layer have to match the size of the input and output vectors, respectively. Each layer can consist of a different number of neurons, where each neuron is connected to all neurons in the adjacent layers. Each connection is characterized by a value called weight. The value of a given neuron is a weighted sum of the values of the neuron in the previous layer multiplied by their corresponding weights plus an added coefficient called bias. The resulting value is fed into a nonlinear function, called activation function, that gives the network its ability to approximate nonlinear latent functions. In regression problems, the main goal of a learning model is to approximate a latent unknown function $f$ given training-set $\mathcal{D}_{tr} = \{\mathbf{u}_i(t), \tau_{GT,i}\}, i = 1, ..., N_{tr}$:

$$\mathcal{NN}(\mathbf{u}(t)|\boldsymbol{\theta}_{\mathcal{NN}}(\mathcal{D}_{tr})) \approx f(\mathbf{u}(t)) \qquad (14)$$

Where $\mathbf{u}(t)$ is the input vector, $\mathcal{NN}$ is the approximation obtained by the learning model, and $\boldsymbol{\theta}_{\mathcal{NN}}$ is the vector of model parameters, i.e. the set of all weights and biases, which are optimized during the learning/training process. During this process, the model parameters are changed to minimize a loss function which is defined *a-priori* as a distance metric between the ground-truth (GT) and predicted outputs of the model. Model parameters are iteratively modified using the Backpropagation algorithm [36]. After the training process is complete, output prediction is obtained by applying the resulting regression function on a new input vector $\mathcal{NN}(\mathbf{u}(t)|\boldsymbol{\theta}_{\mathcal{NN}})$.

As mentioned in the previous section, in the current work we aim not only to find a predicted FT but also a CI concerning a desired CL $\alpha$. The limiting cases in which the CL equals to zero and unity correspond to absolute uncertainty and absence of CI, and perfect certainty with infinitely wide CI, respectively. We use the output of the trained FC-ANN obtained according to Eq. (15) as a reference for the second module of the suggested hybrid model that will be described in the next section. The reference prediction is obtained by the following expression:

$$\eta(\mathbf{u}(t)|\mathcal{D}_{tr}) = \mathcal{NN}(\mathbf{u}(t)|\hat{\boldsymbol{\theta}}_{\mathcal{NN}}(\mathcal{D}_{tr})) \tag{15}$$

Here, $\hat{\boldsymbol{\theta}}_{\mathcal{NN}}$ is the vector of model parameters obtained after the training process, and $\mathcal{D}_{tr}$ is the sets of feature vectors and GT FTs from the train-set. The vector of model parameters $\boldsymbol{\theta}_{\mathcal{NN}}$ is tuned with respect to the following loss function:

$$\begin{aligned} L(\boldsymbol{\theta}_{\mathcal{NN}}|\mathcal{D}_{tr}) &= \mathbb{E}[(\tau_{GT} - \mathcal{NN}(\mathbf{u}(t)|\boldsymbol{\theta}_{\mathcal{NN}}))^2] + \lambda||\boldsymbol{\theta}_{\mathcal{NN}}||_2^2 \\ \hat{\boldsymbol{\theta}}_{\mathcal{NN}}(\mathcal{D}_{tr}) &= \underset{\boldsymbol{\theta}_{\mathcal{NN}}}{\arg\min}\, \mathcal{L}(\mathcal{D}_{tr}) \end{aligned} \tag{16}$$

Here $\lambda$ is the regularization coefficient. The regularization term in the loss function is added to avoid over-fitting. The architecture of the FC-ANN is designed iteratively when the complexity of the network is gradually increased until sufficient accuracy is achieved. The main goal of this stage is to find the "simplest" network that yields good predictions to avoid over-fitting and time-consuming training process. Thus, we obtained a three-layers FC-ANN, with five neurons in the input layer (the size of the reduced input vector $\mathbf{u}(t)$), twelve neurons in the hidden layer, and a single neuron in the output layer. Training is performed using Back-propagation, and mini-batch optimization with the Adam method [37] with an adaptive learning rate with an initial learning rate $\alpha_l = 0.0001$. The model weights are initialized using the Xavier method [38]. To avoid over-fitting, the loss of the model is evaluated on both the train set and a distinct validation set. The training process is completed when the training error had converged or when the maximum number of 500 epochs is reached. The prediction performances of the network are evaluated on the CV-set and compared to the GT results in Fig. 3.

Here, each prediction is described using a single point in the plane. As one can see, the overall performance of the network is described by the distribution of the resulting points around the line of perfect prediction $\tau_{GT} = \eta$. The prediction points are tightly distributed around the line of perfect prediction, indicating that the network successfully captures the underlying relation between the FT and the features, yielding a good surrogate predictive model. However, as expected, the distribution around the line is not negligible and stems from multiple sources of stochasticity and uncertainty in measurement methodology, values of system parameters, etc. Hence, a probabilistic module than can quantify uncertainty associated with the FC-ANN FT predictions is essential for our model.

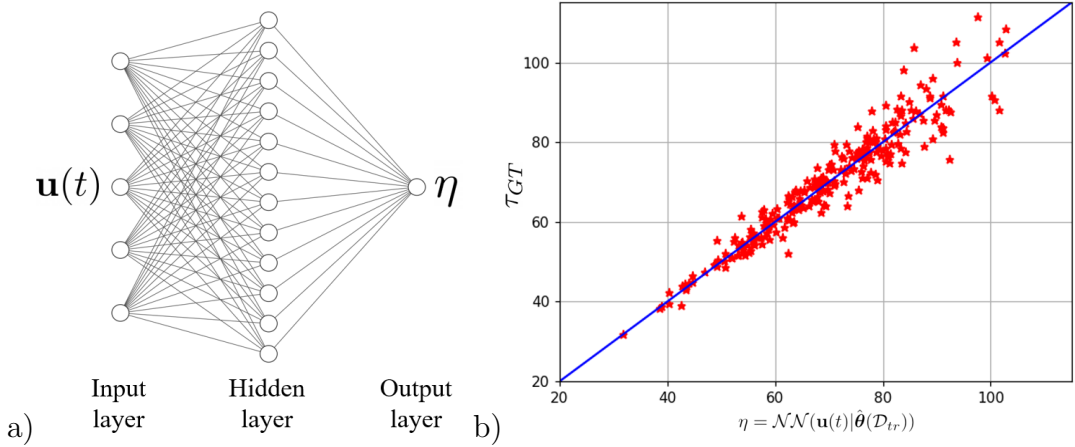

Figure 3: a) Scheme of fully-connected artificial neural network (FC-ANN) with a single hidden layer that learn the mapping between input feature vector $\mathbf{u}(t)$ and the corresponding predicted FT with respect to its model parameters: $\eta = \mathcal{NN}(\mathbf{u}_i|\hat{\boldsymbol{\theta}}_{\mathcal{NN}}(\mathcal{D}_{tr}))$. The predicted FT $\eta$, together with the GT FT $\tau_{pred}$, serves as a representation of the input vector $\mathbf{u}(t)$ in a reduced-order embedding space; b) representation of the input vector $\mathbf{u}(t)$ in a reduced-order embedding space which shows a comparison between GT and predicted results obtained by the trained FC-ANN $\eta$ (red dots), and line of perfect prediction (blue line).

The variance of the results obtained for each value of $\eta$ embodies the level of uncertainty associated with the ANN prediction. Moreover, the variance of the distribution of the results varies with $\eta$. Hence, we treat the prediction points as representations of the features vector in a reduced-order embedding space $\eta - \tau_{GT}$. We aim to fit a multivariate PDF above the embedding space that maps each prediction $\eta(\mathbf{u}(t))$ to a PDF that simultaneously embodies the most likely FT value and corresponding prediction uncertainty. The level of uncertainty will be dictated by the amount of similar data in the training-set, i.e. near points in the embedding space. For highly exploited regions of $\eta$ the certainty level is high, and we expect to get lower variances, and vice versa. In the next section, we introduce a GPR-based approach for UQ of the predictive results obtained by the FC-ANN for a real-time measured features vector $\mathbf{u}(t)$.

### 3.2. Gaussian process regression

Gaussian process regressions (GPRs) [28] are non-parametric kernel-based probabilistic models that take as input training dataset $\mathcal{D}_{\text{CV}} = \{\eta_i, \tau_i\}, i = 1, \ldots, N_{\text{CV}}$ of $N_{\text{CV}}$ pairs of input variables $\boldsymbol{\eta}_i$ and noisy scalar value $\tau_i$, and constructs a surrogate model that generalizes well for unseen data using a multivariate PDF. The PDF is used for simultaneous prediction of the output value and the corresponding uncertainty. The noise in the output variable $\tau_i$ models uncertainty due to various factors such as a systematic error in measuring the system properties and stochastic measured signal. We define a latent function $g(\tau)$ that maps ANN predictions $\eta$ to the corresponding GT TFs $\tau_{pred}$, i.e $g(\eta) : \eta \mapsto \tau_{GT}$ Here, we assume that the latent function $g(\eta)$ is inaccessible, and can be evaluated only using noisy observations $\tau$. Moreover, we assume that the noise is additive zero-mean, and normally distributed.

$$\tau = g(\eta) + \epsilon, \epsilon \sim \mathcal{N}(0, \sigma_n^2) \tag{17}$$

Here, $\sigma_n^2$ is the variance of the noise. A Gaussian process (GP) is a collection of random variables, any finite number of which have a joint Gaussian distribution. The main idea behind GPR is to utilize a GP to represent the latent function $g(\eta)$. This is done in a Bayesian framework, in which the GP serves prior over function-space which in turn is conditioned on the measured data to yield a posterior distribution. The latter enables us to make an inference on the values of the latent function $g(\tau)$ that correspond to unseen input variables $\eta$ using a finite set of training data $\mathcal{D}_{\text{CV}}$.

$$g(\eta) \sim \mathcal{GP}(\mu_0(\eta), \Sigma_0(\eta, \eta')) \tag{18}$$

A GP is completely specified by its mean function and covariance function, $\mu_0(\eta)$ and $\Sigma_0(\eta, \eta')$, respectively.

$$\mu_0(\eta) = \mathbb{E}[g(\eta)], \quad \Sigma_0(\eta, \eta') = \mathbb{E}[g(\eta - \mu_0(\eta))g(\eta' - \mu_0(\eta'))] \tag{19}$$

Here $\mathbb{E}$ denotes expectation. The mean functions $\mu_0(\eta)$ of the prior GP embodied a prior knowledge about the latent function $g(\tau)$. Usually, it is unknown and therefore set to zero. The properties of the GP are governed by the covariance function or kernel, which is symmetric and positive semi-definite by definition. The covariance function $\Sigma_0(\eta, \eta')$ quantifies the covariance between any pair of points in the dataset with respect to a given covariance function, also called kernel. The most commonly used is the squared exponential kernel, also known as the Radial Basis Function (RBF) kernel, or Gaussian kernel.

$$\Sigma_0(\eta, \eta') = \sigma_l^2 \exp\left(-\frac{(\eta - \eta')^2}{2l^2}\right) \tag{20}$$

Here parameter $l$ is a characteristic length scale that repents the covariance between a data point to its neighbors and thus is dictating the smoothness of the candidate functions generated by the GP. For example, large values of $l$, enforce large off-diagonal values in the covariance matrix, which corresponds to smooth candidate function, and vice versa. In general, extrapolation using GPR is reliable only approximately $l$ units away from the training data. Parameter $\sigma_l$ determines the variance of the candidate functions away from the mean $\mu_0(\eta)$. The set of parameters $\boldsymbol{\theta}_{\text{GPR}} = \{\sigma_n, l, \sigma_l\}$, also called hyper-parameters, are are determined to best fit the data by optimizing over log-likelihood loss function according to the maximum *a-posteriori* (MAP) principle:

$$\mathcal{L}_{\text{MAP}}(\boldsymbol{\theta}_{\text{GPR}}) = log(\tau|\boldsymbol{\theta}_{\text{GPR}}) = \frac{1}{2}\boldsymbol{\tau}^T(K_{\boldsymbol{\tau},\boldsymbol{\tau}})^{-1}\boldsymbol{\tau} - \frac{1}{2}\log K_{\boldsymbol{\tau},\boldsymbol{\tau}} - \frac{1}{2}N_{\text{CV}}\log 2\pi \tag{21}$$

Here, $K$ is the symmetric covariance matrix whose $ij^{th}$ entry is the covariance between the $i^{th}$ variable in the group denoted by the first subscript and the $j^{th}$ variable in the group denoted by the second subscript, calculated using covariance function $\Sigma_0$ and corresponding hyper-parameters. Vector $\boldsymbol{\tau}$ is a vector of the training observations, and $K_{\boldsymbol{\tau},\boldsymbol{\tau}} \equiv K_{\mathbf{g},\mathbf{g}} + \sigma_n \mathbb{I}$. The vector of optimal hyper-parameters $\hat{\boldsymbol{\theta}}_{\text{GPR}}$ is obtained by optimization the log-likelihood loss function over the hyper parameter space $\boldsymbol{\Theta}_{\text{GPR}}$:

$$\hat{\boldsymbol{\theta}}_{\text{GPR}} = \underset{\boldsymbol{\theta}_{\text{GPR}} \in \boldsymbol{\Theta}_{\text{GPR}}}{\arg\min} \mathcal{L}_{\text{MAP}}(\boldsymbol{\theta}_{\text{GPR}}) \tag{22}$$

Differentiating Eq. (22) with respect to the hyper-parameters yields:

$$\frac{\partial \mathcal{L}_{\text{MAP}}}{\partial \theta_{GPR,i}} = \frac{1}{2}\text{Tr}\left(\left(K_{\tau,\tau}^{-1}\boldsymbol{\tau}\left(K_{\tau,\tau}^{-1}\boldsymbol{\tau}\right)^T - K_{\tau,\tau}^{-1}\right)\frac{\partial K_{\tau,\tau}}{\partial \theta_{GPR,i}}\right) \qquad (23)$$

The computational complexity of this calculation is dictated by the matrix inverse, which is $\mathcal{O}(N_{\text{CV}}^3)$. After the optimal hyper-parameters $\hat{\boldsymbol{\theta}}_{\text{GPR}}$ over the training-data are obtained, the GPR is fully-determined and inference can be easily made by computing the posterior PDF obtained after conditioning on the training-data $\mathcal{D}_{\text{GPR}} = \{\boldsymbol{\eta}, \boldsymbol{\tau}_{GT}\}$.

$$\boldsymbol{\tau}^*|\mathcal{D}_{\text{GPR}}, \boldsymbol{\eta}^* \sim \mathcal{N}(\boldsymbol{\mu}(\boldsymbol{\eta}^*|\mathcal{D}_{\text{GPR}}), \boldsymbol{\Sigma}(\boldsymbol{\eta}^*|\mathcal{D}_{\text{GPR}})) \qquad (24)$$

Here, $\boldsymbol{\tau}$ is the estimated observations vector that corresponds to the vector of test inputs $\boldsymbol{\eta}$, for which we want to estimate the latent function $g(\eta)$. $\boldsymbol{\mu}(\boldsymbol{\eta}^*|\mathcal{D}_{\text{GPR}})$ and $\boldsymbol{\Sigma}(\boldsymbol{\eta}^*|\mathcal{D}_{\text{GPR}}))$ are the mean and covariance function of the posterior GP, obtained after conditioning the prior of the training data $\mathcal{D}_{\text{GPR}}$.

$$\begin{aligned}\boldsymbol{\mu}(\boldsymbol{\eta}^*|\mathcal{D}_{\text{GPR}}) &\equiv \mathbb{E}\left[\boldsymbol{\tau}^*|\mathcal{D}_{\text{GPR}}, \boldsymbol{\eta}^*\right] = K(\boldsymbol{\tau}^*, \boldsymbol{\tau})\left(K(\boldsymbol{\tau},\boldsymbol{\tau})+\sigma_n^2\mathbb{I}\right)^{-1}\boldsymbol{\tau}_{GT} \\ \boldsymbol{\Sigma}(\boldsymbol{\eta}^*|\mathcal{D}_{\text{GPR}}) &= K(\boldsymbol{\eta}^*,\boldsymbol{\eta}^*) - K(\boldsymbol{\eta}^*,\boldsymbol{\eta})\left(K(\boldsymbol{\eta},\boldsymbol{\eta})+\sigma_n^2\mathbb{I}\right)^{-1}K(\boldsymbol{\eta},\boldsymbol{\eta}^*)\end{aligned} \qquad (25)$$

One of the main advantages of GPs, is that conditioning the resulting multivariate-PDF on a given input variable yields a Gaussian posterior univariate PDF on the output variable. Hence, the posterior GP is now used to generate a Gaussian PDF over the space of FTs for a given input value $\eta$ that corresponds to an input features vector $\mathbf{u}(t)$, as shown in Eq. (26)

$$\tau(\mathbf{u}(t)|\boldsymbol{\theta}) \sim f_{pr}(\tau|\mathbf{u}(t),\boldsymbol{\theta}) = \mathcal{N}\left(\mu\left(\eta(\mathbf{u}(t)|\hat{\boldsymbol{\theta}}_{\mathcal{NN}})|\hat{\boldsymbol{\theta}}_{\text{GPR}}\right), \Sigma\left(\eta(\mathbf{u}(t)|\hat{\boldsymbol{\theta}}_{\mathcal{NN}})|\hat{\boldsymbol{\theta}}_{\text{GPR}}\right)\right) \qquad (26)$$

Here, $\mu$ and $\Sigma$ are the mean and variance of the resulting Gaussian PDF $f_{pr}$, respectively. The standard deviation of the PDF is defined as follows: $\sigma(\eta) = \sqrt{\Sigma(\eta)}$. Parameters vector $\boldsymbol{\theta}$ includes the parameters of both modules of the hybrid model, i.e. $\boldsymbol{\theta} = \{\hat{\boldsymbol{\theta}}_{\mathcal{NN}}(\mathcal{D}_{tr}), \hat{\boldsymbol{\theta}}_{\text{GPR}}(\mathcal{D}_{\text{CV}})\}$. The PDF $f_{pr}(\tau|\mathbf{u}(t),\boldsymbol{\theta})$ is calculated in real-time for each measured features vector $\mathbf{u}(t)$.

The last information to be fused into the PDF over FTs is the measurement instance $t$, by eliminating the probability for failure before this instance by introducing a posterior PDF for which $f_{post}(\tau < t) = 0$. This will allow enhanced confidence for relevant FTs and shrinking the CI associated with the failure prediction. Thus, we introduce the following Heaviside likelihood function and calculate the posterior PDF over the FTs using the Bayes rule:

$$\begin{aligned}\mathcal{L}(\tau|t) &= H(\tau - t) \\ f_{post}(\tau|\mathbf{u}(t),\boldsymbol{\theta},t) &= \frac{f_{pr}(\tau|\mathbf{u}(t),\boldsymbol{\theta})\mathcal{L}(\tau|t)}{\int_{-\infty}^{\infty}f_{pr}(\tau'|\mathbf{u}(t),\boldsymbol{\theta})\mathcal{L}(\tau'|t)d\tau'}\end{aligned} \qquad (27)$$

Now, FT prediction and the estimation of the corresponding CI are intuitive. According to the maximum *a-posteriori* probability (MAP) principle, the predicted FT is estimated as the mode of the resulting posterior distribution. The boundaries of the CI, denoted by $\tau^-, \tau^+$, correspond to cumulative distribution of $\alpha$ with respect to the posterior PDF as shown in Eq. (28), and by that enforce the desired CL.

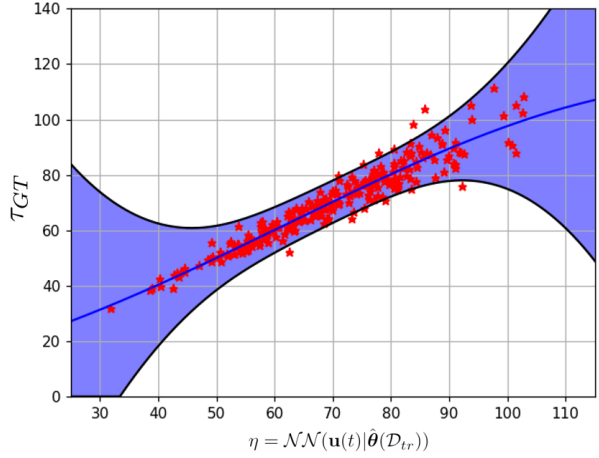

Figure 4: True failure times $\tau_{GT}$ vs. the predictions made by the pre-trained FC-ANN (red dots), the mean function of the posterior GP $\mu(\eta(\mathbf{u}(t)))$, CI corresponding to CL of $\alpha = 95\%$ and two standard deviations $\sigma(\eta) = \sqrt{\Sigma(\eta(\mathbf{u}(t))}$ with respect to the mean (shaded blue), and the CI boundaries (solid black).

$$\tau_{pred} = \arg\max_{\tau} f_{post}(\tau|\mathbf{u}(t), \boldsymbol{\theta}, t)$$
$$\int_{\tau^-}^{\tau^+} f_{post}(\tau)d\tau = \alpha \qquad (28)$$

The variance of the posterior PDF is a measure of the uncertainty associated with the prediction obtained. More specifically, since the posterior PDF is Gaussian or nearly-Gaussian, the boundaries of the CI can now be obtained analytically for a desired CL $\alpha$ as follows:

$$\begin{aligned}
\tau^-, \tau^+(\alpha|\mathbf{u}(t), \boldsymbol{\theta}, t) &= \begin{cases} \tau_{pred} \pm \gamma(\alpha)\sigma & , t \in (0, \tau_{pred} - \gamma(\alpha)\sigma] \\ t, \tau_{pred} + \nu(\alpha)\sigma & , \text{else} \end{cases} \\
\gamma(\alpha|\mathbf{u}(t), \boldsymbol{\theta}) &= \sqrt{2}\mathrm{erf}^{-1}\left(\frac{\alpha}{2}(1+\kappa)\right) \\
\nu(\alpha|\mathbf{u}(t), \boldsymbol{\theta}, t) &= \sqrt{2}\mathrm{erf}^{-1}\left(\alpha(1+\kappa) - \kappa\right) \\
\sigma &= \sqrt{\Sigma(\eta(\mathbf{u}(t)))}, \kappa = \mathrm{erf}\left(\frac{\tau_{pred} - t}{\sqrt{2}\sigma}\right)
\end{aligned} \qquad (29)$$

As one can see in Eq. (29), the penetration of the measurement time into the CI of the prior PDF, leads to asymmetricity in the CI with respect to the mode of the posterior PDF. Moreover, as the measurement time advances and more knowledge is accumulated regarding the probable FT (more precisely when the fatigue failure does not occur), the CI shrinks as the level of uncertainty decreases. This effect is demonstrated in Fig. 5 for a simplified case of constant predicted FT $\tau_{pred}$.

As one can see in Fig. 5, due to constant fusion between the prior PDF and measured observations, the prediction certainty level increases with time as the size of the CI decreases. Generally, the predicted FT can oscillate erratically due to the stochastic nature of the signal and its estimated statistical features in the features vector $\mathbf{u}(t)$. In this case, the size of the CI oscillates as well and is dramatically influences by the exploitation levels of the resulting FC-ANN prediction $\eta$, i.e. high standard deviation are obtained in unexploited regions of $\eta$, and vice versa. However, as the measurement time approaches the predicted FT $\tau_{pred}$, the Bayesian inference regarding the FT (Eq. (27)) promises an extensive increase in the degree of certainty of the resulting prediction. In summary, the suggested hybrid model maps between a real-time measured features vector to a PDF that is used for both FT prediction and UQ with respect to a given CL $\alpha$ and in a measurement frequency $f_s$ of request. In the next section, the real-time performances of the hybrid model are demonstrated for an unseen stochastic signal.

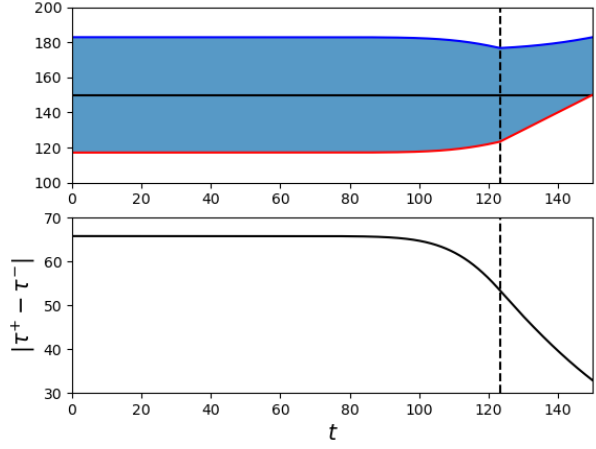

Figure 5: Illustration of the CI vs. time for a constant predicted FT $\tau_{pred} = 150$ (solid black) for CL $\alpha = 90\%$ and prior standard deviation of $\sigma = 20$. Top: the CI vs. time (shaded blue), upper and lower CI boundaries (solid blue and red, respectively), the instance in which the measurement time equals to the lower CI boundary, i.e. $t = \tau_{pred} - \gamma(\alpha)\sigma$ (dashed black). Bottom: CI size vs. time (solid black).

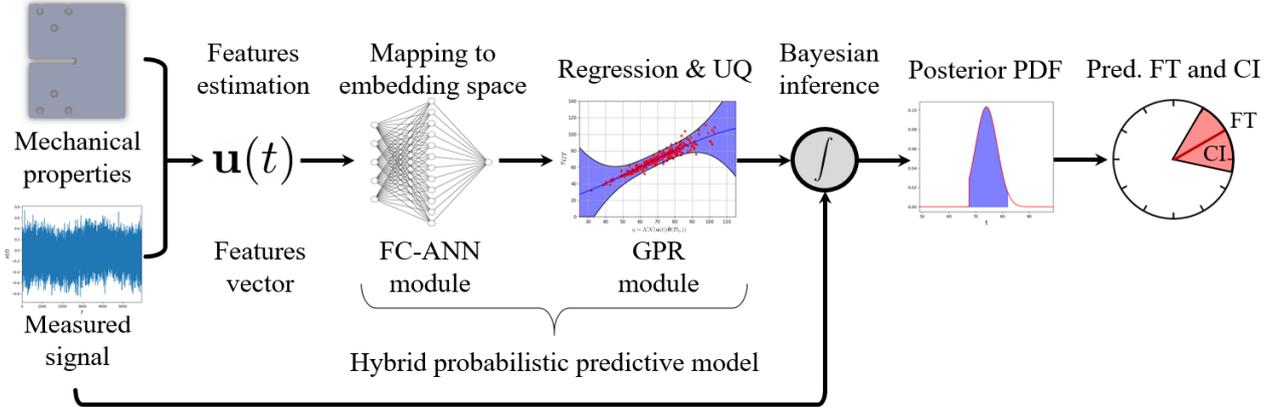

Figure 6: Schematic description of the hybrid predictive model, including data flow and interactions between the different modules.

## 4. Real-time fatigue prediction and uncertainty quantification

After both modules were trained on the train-set (FC-ANN) and cross-validation-set (GPR), the prediction and UQ performances of the hybrid predictive model can be demonstrated on an unseen stochastic signal taken from the test set. To simulate a real-time framework, the signal is sampled at a constant rate of $f_s = 1$Hz. In each sampling, the measured signal $x(t)$ is used to accumulate the temporal features vector $\mathbf{u}(t)$ according to Eq. (4), (11). We assume that the material fatigue properties are known with finite accuracy. The measured stochastic signals $x(t) \sim X(t,\zeta)$ can be chosen as any physical observable that is strongly correlated with the stress in the critical section on the mechanical component which is the most prone to fatigue failure. For simplicity and explainability, in the current illustration, all signals are considered as stress signals measured directly from the critical section. All signals in the dataset are generated from the same PSD. Thus, the modified features vector is used. Otherwise, the moments of the PSD should have been calculated using Welch's method and included in the extended features vector. Then, the pre-trained FC-ANN is used for generating a prior prediction of the FT $\eta$ according to Eq. (15). The GPR is used for calculating the posterior PDF $f_{post}(\tau)$ according to Eq. (17)-(28). Finally, the predicted FT and CI boundaries are calculated using the posterior for a desire CL $\alpha$. The results are shown in Fig. 7 for four representative time instances. As one can see, approximately after half of the signal was measured, the GT FT falls within the CI. As measurement time approaches the GT FT, the CI becomes narrower - indicating a higher level of certainty achieved during the learning process due to constant evolution of the posterior PDF.

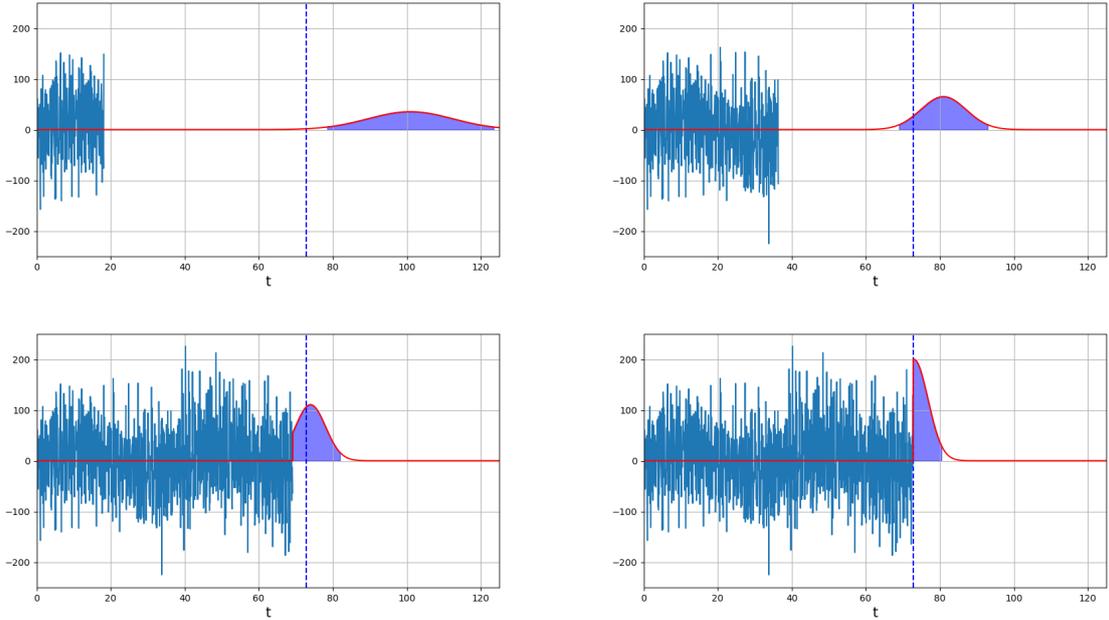

Figure 7: Demonstration of real-time FT prediction, in four consecutive instances corresponding to $t = [25\%, 50\%, 95\%, 100\%]\,\tau_{GT}$, respectively. The measured loading signal (solid blue), GT FT $\tau_{GT}$ (dashed blue), posterior PDF of FTs (solid red), and CI corresponding to CL of $\alpha = 95\%$ (purple shade). The predicted FT $\tau_{pred}$ and the confidence boundaries $\tau^-, \tau^+$ correspond to the mode of the posterior PDF and boundaries of the CI, respectively.

## 5. Experimental validation

The hybrid predictive model is validated using experimental data obtained by the Society of Automotive Engineers (SAE) [19]. This data was collected is to serve as a benchmark for fatigue prediction methods for ground vehicles. The examples in the dataset include keyhole-shaped steel specimens made of US Steel's Man-Ten alloys and Bethlehem's RQC-100. The properties of both materials are shown in Table 2. The specimens are of identical geometry and are hot-rolled in parallel to the direction of crack growth (perpendicular to the applied load).

### 5.1. Finite element analysis for stress concentration factor estimation

Due to the existence of the notch, stress concentration effects become significant and magnify the lead to substantial magnification of the stresses in the vicinity of the notch. This effect is considerd by stress concentration factor $K_t$ which is defined as the relation between the maximal actual stress measured near the notch $\sigma_{max}$ and the nominal stress $\sigma_{nom}$, as

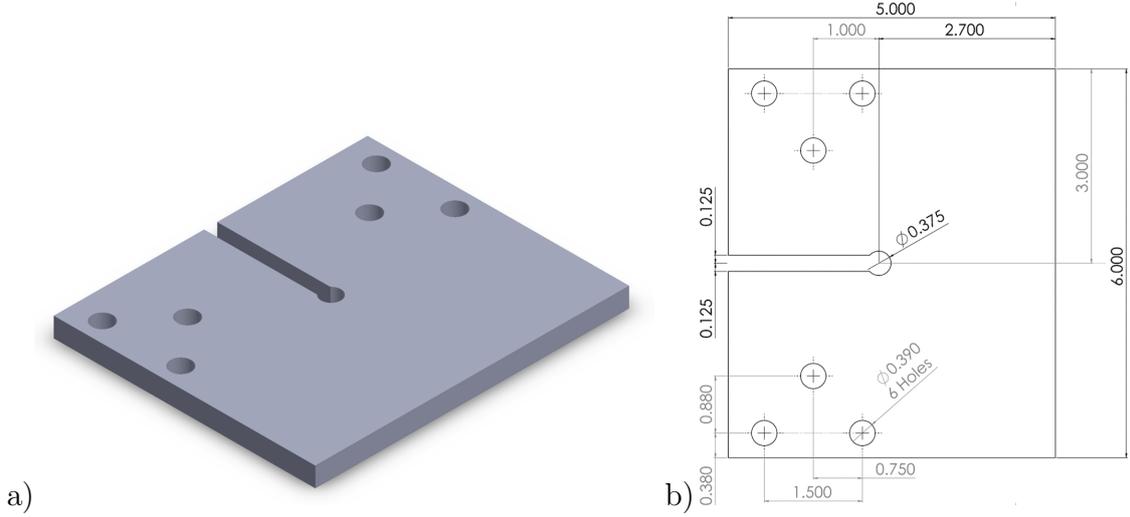

Figure 8: Notched keyhole experimental specimen; a) isometric view, b) top view drawing, all dimensions are in inches.

| Property | Man-Ten | RQC-100 |
|---|---|---|
| Modulus of Elasticity, E, GPa | 203 | 203 |
| Yield Strength, $\sigma_y$, MPa | 327 | 586 |
| Ultimate Tensile Strength, $\sigma_{uts}$, MPa | 565 | 820 |
| Fatigue Strength Coefficient, A, MPa | 917 | 1160 |
| Fatigue Strength Exponent, b | -0.095 | -0.075 |

Table 2: Fatigue properties of the experimental specimens, [19].

shown in Eq. (30):

$$K_t = \frac{\sigma_{max}}{\sigma_{nom}} \tag{30}$$

There is a substantial variance among estimation of the stress concentration factor in keyhole specimens reported in the literature, that range between 2.62 to 3.52 [39, 40]. Moreover, most of those studies use outdated methods for evaluating factor $K_t$. Hence, we conduct a dedicated 3D finite element analysis (FEA) for evaluating the resulting stress concentration factor of the specimen. The FEA was performed using the commercial package SolidWorks Simulation 2020. As the keyhole specimen is symmetric in the longitudinal direction the analysis was performed for half of the specimen only with symmetry boundary condition applied accordingly. Equal radial loads were applied on each of the top halves of the grip holes to emulate the load pattern of the clevis pins. Loads of $2,965N$ were distributed sinusoidally in the circumferential direction. Moreover, zero in-plane translation restraints were applied on the faces of the grip holes. The mesh of the finite element model was generated using parabolic tetrahedral elements with a global element length of $2.00mm$, as shown in Fig. 9. A refined mesh was generated in the vicinity of the notch with a length of $0.1mm$ and a maximum aspect ratio of 1.2. This resulted in a total of $433,235$ nodes and $295,675$ elements. The circumferential stress distribution $\sigma_{theta}$ in the notch vicinity is presented in Fig. 10.

As one can see in Fig. 10, the maximum circumferential stress is obtained at the mid-plane and equals 312.6MPa. The nominal stress is calculated using Eq. (31).

$$\sigma_{nom} = \frac{P_e}{A_s} + \frac{M_e z}{I} \tag{31}$$

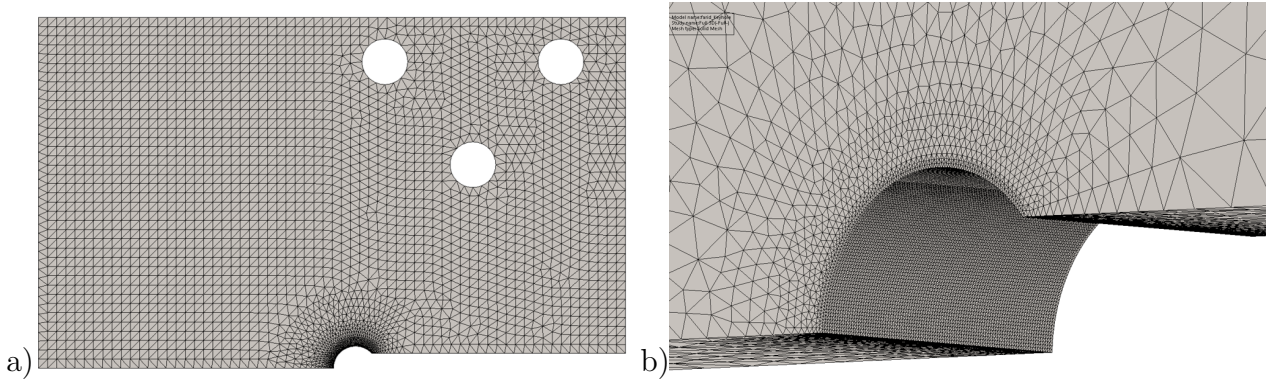

Figure 9: Mesh of the 3D finite element model, a) half of the model about the plane of symmetry, b) Zoom-in in the vicinity of the specimen's notch.

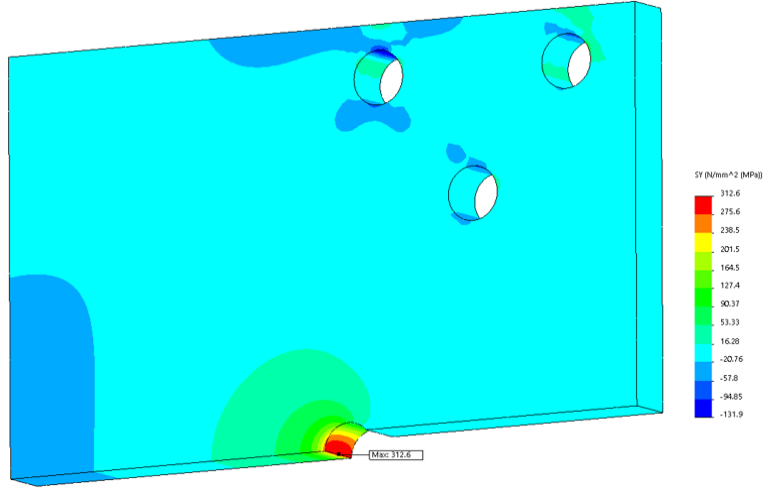

Figure 10: Stress distribution on the 3D FE model and the resulting stress concentration near the notch.

Where $P_e$ and $M_e$ are external force and moment applied on the specimen, respectively. Parameters $A$ and $I$ are the area and second moment of area of the critical section, respectively. Variable $z$ is the coordinate along the axis of symmetry. According to Eq. (31), for pure tension load of $P_{ex} = 8.895$kN, resulting nominal stress of 100.2MPa is obtained. Employing the relations shown in Eq. (31)-(30), the resulting stress concentration factor obtained is $K_t = 3.12$. This value will be used for further calculation of the equivalent stresses according to Eq. (30), i.e. $\sigma_{max} = K_t \sigma_{nom}$.

### 5.2. Validation of the hybrid predictive model using experimental data

For all notched specimen tests, loads were applied to the specimen through a close tolerance monoball fixture which allowed both tensile and compressive loading. The loading signals are given as series of peaks and valleys, disregarding the frequency content of the signals (nominally between $1-30$Hz). For scaling purposes, all signals are normalized to fit the range of $\pm 1$. The dataset includes three types of loading signals which differ in their statistical properties: a) A load history obtained from the bending moment on a vehicle suspension component, driven over an accelerated durability course, b) a vibration obtained on a mounting bracket being excited by vehicle operation over a rough road, c) A load history with a drastic change of a mean load obtained from transmission torque measured on a tractor engaged in front-end loader work. In each loading regime, the actual loading signals are obtained by a linear scaling of the normalized signals with respect to three load levels: 15.6kN, 35.6kN, and 71.2kN. The three variable amplitude histories are linearly scales to three levels with three replications of each level and material to produce a variety of FTs. For each experiment, three characteristic instances were recorded: crack initiation, crack propagation, and total life until failure which is the sum of the first two. The SAE dataset includes 57 samples in total. In the current analysis, we consider the time histories that correspond to the second loading type, i.e. 20 samples in total (Fig. 11).

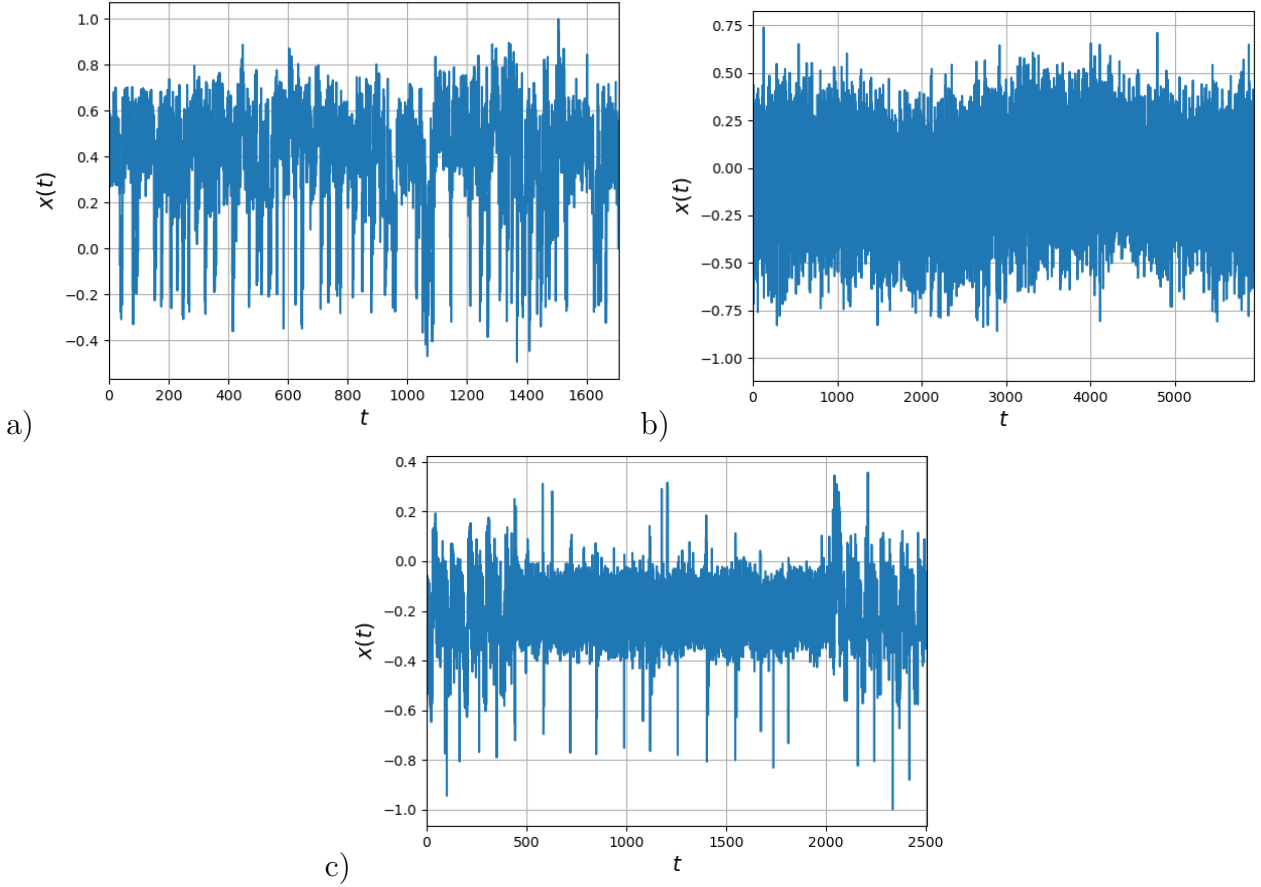

Figure 11: Normalized time histories of distinct vehicle component during driving in rough road taken from SAE dataset; a) bending moment on a vehicle suspension component, b)displacement of bracket mounting, c) transmission torque measured on a tractor engaged in front-end loader work.

To simulate real-time sensing, the peaks and valleys sequence is interpolated using cubic spline interpolation in the frequency of $10 f_{max}$ where $f_{max} = 35$Hz. As mentioned in previous sections, training of hybrid predictive model, approximately $N = 1500$ samples are required. Hence, additional 1480 stochastic loading signals are augmented and generated from the underlying PSD associated with the second loading type. The latter is obtained using Welch's method for PSD estimation from a time-history, as explained in Section 2. Signals are generated according to Eq. (2) and $N_f = 20$ components and frequency range of $f \in (0, 35]$Hz. All signals are multiplied by the stress concentration factor $K_t$ to yield the equivalent stress in the vicinity of the notch of the keyhole specimen. FTs of the augmented stochastic loading signals are estimated using rainflow counting and Miner's rule. Both the augmented and the authentic dataset are being split into training, CV, and test sets with a ratio of 70 : 15 : 15, respectively. The first and the second sets are combined and shuffled and are uses to train the FC-ANN and GPR modules. The third set is used to evaluate the performances of the trained hybrid model on the augmented and authentic data separately. To evaluate the model's real-time prediction performances, we introduce a success criterion that simultaneously describes prediction accuracy, earliness, and stability. A Prediction is considered as accurate if GT FT falls within the CI of the posterior PDF at $\beta$ percent of the time until failure for $r$ percent of the time.

| Dataset | Accuracy | Inaccurate predictions | |
|---------|----------|------------------------|---|
| | | Conservative pred. | Non-conservative pred. |
| Total[1] test-set | 88.2%(1323) | 62.71%(111) | 37.29%(66) |
| Authentic test-set | 83.3%(5) | 100%(1) | – |

Table 3: Summary of the results obtained by the hybrid probabilistic predictive model in the SAE experimental dataset [19] on the total dataset (both authentic and augmented data) and on the authentic data for $\beta = 60\%$ and $r = 75\%$. The performances of the model are describes by prediction accuracy and classification of the model inaccuracies to conservative prediction and underestimations. The former corresponds to $\tau_{GT} > \tau^+$, and the latter to $\tau_{GT} < \tau^-$.

$$\tau_{GT} \in [\tau^-, \tau^+](\alpha, \mathbf{u}(t \in [r, 1]\beta\tau_{GT}, \boldsymbol{\theta}, t = \beta\tau_{GT}) \tag{32}$$

Here, $\beta$ and $r$ are user-defined parameters that control the level of early prognosis and prediction stability, respectively. In the current work, we take the early prediction and stability parameters as $\beta = 60\%$ and $r = 75\%$, respectively. The CL is chosen as $\alpha = 95\%$ The results are shown in Table 3. As one can see, the hybrid model captured approximately 89% of FTs. In the other cases, it tends to produce a conservative estimation, i.e. $\tau^+ < \tau_{GT}$, rather than an underestimation, i.e. $\tau^- > \tau_{GT}$. However, this is not a built-in feature of the suggested model, but a statistical property that derives from the PSD of the signal, and the occurrence frequency of extreme amplitudes that cause drastic changes in the measured statistical featured of the signal. The results improved significantly when favoring accuracy over early prediction, i.e. as the early prediction parameter $\beta$ is closer to unity, and allowing the features vector to "stabilize" on its steady-state values. The inaccurate predictions obtained on the authentic dataset cannot be considered representative due to the small number of samples in this dataset (six). Better performances can be obtained by optimizing the architecture of the FC-ANN modules and increasing the size of the dataset to 5000 samples and above.

As one can see in Table 3, the FT predictions obtained by the hybrid model are promising and in good agreement with the experimental data for a variety of material parameters, loading intensities, and FTs. For 88.2% of examples in the total dataset, the FTs were captured withing the CI in time duration of $(1 - \beta)\tau_{pred}$ before failure occurrence and with prediction stability of $r$. Similar results were obtained for the authentic experimental dataset. When the model obtained inaccurate predictions, most of them were conservative (62.71%), i.e. the GT FT was larger the upper boundary of the CI, $\tau^+ < \tau_{GT}$. Compensation on less early predictions, i.e. larger values of $\beta$, allows the estimated statistical properties of the signal $\bar{x}$ and $\xi(\rho)$ to reach steady-state and to base the posterior PDF on richer data. Thus, the model predictions improve for larger values of parameter $\beta$.

In contradiction to most of the existing methods, this method allows real-time failure prediction thanks to a real-time features estimation methodology. The usage of confidence interval with an adaptive size, i.e. confidence level, allows obtaining upper and lower boundaries for the predicted FT, which in turn, serves a major role in real-time decision making in terms of safe operation, maintenance, and structural health monitoring. Future research will deal with data-driven probabilistic fatigue prediction for non-stationary stochastic signals.

## 6. Concluding remarks

A hybrid data-driven method for real-time prediction and uncertainty quantification (UQ) of fatigue failure under stochastic loading was formulated. The developed approach combines two powerful learning models, fully-connected artificial neural networks (FC-ANNs) and Gaussian Process Regression (GPR), to answer an essential engineering need of real-time prediction of time *intervals* that are prone to fatigue failures with a given certainty level (CL). The proven ability of the FC-ANN to represent high-dimensional data in a reduced-order embedding space is utilized by the GPR for enhanced prediction and simultaneous UQ. After the prior training process of both predictive modules, the predictions are obtained at an extensive speed, making the methodology suitable for real-time purposes. Moreover, a full methodology was suggested for real-time evaluations of the input features calculated based on the measured signal, both for the case in which the signals are generated from a single PSD or multiple PSDs. The suggested method is applicable for stochastic signals of various physical properties, such as stress and strain, that are correlated to the stress in the critical section of the studied mechanical component. The proven prediction and generalization abilities of the FC-ANN to represent the data in a latent embedding space are used by the GPR for simultaneous prediction and uncertain quantification. Moreover, a full methodology was suggested for real-time evaluations of the input features calculated based on the measured signal, both for the case in which the signals are generated from a single PSD or multiple PSDs. The method is applicable for measured data obtained by experiments, numerical analysis, time-domain, or frequency-domain methods. We have demonstrated the real-time performances of the model for an unseen signal, and a posterior updating rate of $f_s = 1$Hz. The FT was accurately predicted in a safety interval of 50% of the signal duration. The suggested hybrid method paves the way towards the development of predictive systems to be used for generating safety and operational instructions in real-time for failure prevention in mechanical systems under stochastic loadings, structural health monitoring (SHM) purposes, and protection from irreversible damage to the environment and human lives.

## Acknowledgments

The author is grateful to Dr. P. Kravets and Dr. S.H. Rudy for enjoyable discussions, and Prof. J.F. Durodola and Prof. M. Groper for sharing valuable experimental and numerical data.

## Funding

The author has been supported by the Fulbright Program, the Israel Scholarship Education Foundation (ISEF), Jean De Gunzburg International Fellowship, the Israel Academy of Sciences and Humanities, the Yitzhak Shamir Postdoctoral Scholarship of the Israeli Ministry of Science and Technology, and the PMRI – Peter Munk Research Institute - Technion.

**List of Abbreviations**

| | |
|---|---|
| CL, CI | Confidence level and confidence interval |
| CV | Cross-validation |
| FC-ANN | Fully-connected artificial neural network |
| FEA | Finite element analysis |
| FT, TtF | Failure time, Time to failure |
| GPR | Gaussian process regression |
| GT | Ground truth |
| MAP | Maximum *a-posteriori* estimation |
| ML | Machine learning |
| PDF | Probability density function |
| PSD | Power spectral density |
| UQ | Uncertainty quantification |

**List of Symbols**

| | |
|---|---|
| $\bar{\alpha}$ | Goodman's correction factor for non-zero mean stress |
| $\beta, r$ | Early prediction parameter and prediction stability parameter |
| $\boldsymbol{\theta} \in \boldsymbol{\Theta}$ | Set of all parameters of the hybrid model, space of model parameters |
| $\boldsymbol{\theta}_{\text{GPR}}$ | Hyper-parameters vector of the GPR, $\boldsymbol{\theta}_{\text{GPR}} = \{\sigma_n, l, \sigma_l\}$ |
| $\Delta f$ | Frequency resolution of the PSD |
| $\eta$ | FC-ANN representation for input vector $\mathbf{u}(t)$ in the embedding space |
| $\mathbf{u}(t)$ | Features/input vector |
| $\mathcal{D}, \mathcal{D}_j$ | Dataset, training, CV and test subsets, $j = tr, CV, test$ |
| $\mathcal{L}(\tau)$ | Likelihood function |
| $\mathcal{NN}(\mathbf{u}; \boldsymbol{\theta}_{\mathcal{NN}})$ | Regression function associated with ANN with model parameters $\boldsymbol{\theta}_{\mathcal{NN}}$ |
| $\mathcal{Q}$ | Operator that maps a signal to the set of its local maxima |
| $\mu, \Sigma, \sigma$ | Mean, variance, and standard deviation of the posterior GP |
| $\mu_0, \Sigma_0, K$ | Mean, variance, and covariance matrix of the prior GP |
| $\mu_G, \sigma_G$ | Mean and standard deviation of the Gaussian-like PSD |
| $\Phi$ | Cumulative distribution function of the standard normal distribution |
| $\rho, \xi(\rho)$ | Percentile and percentile amplitude |
| $\tau, \tau^+, \tau^-(\alpha)$ | FT, Upper and lower CI boundaries for a given CL $\alpha$ |
| $\tau_{pred}, \tau_{GT}$ | Predicted and GT FTs |
| $\tilde{\mu}, \tilde{\sigma}$ | Mean and standard deviation of the set of local maxima |
| $A, b, \sigma_{uts}$ | Fatigue strength, fatigue exponent, and ultimate tensile strength |
| $D(t)$ | Cumulative damage fraction |
| $f_i, \phi_i$ | The frequency and phase of the $i^{th}$ term in $\chi(t)$ |
| $f_s$ | Sampling/prediction rate |
| $f_{pr}, f_{post}$ | Prior and posterior PDFs |
| $g(\eta), \epsilon$ | Latent function that maps ANN predictions to GT FTs, zero-mean Gaussian noise |
| $G(f)$ | PSD function |
| $H(\tau)$ | Heaviside step function |
| $k_s$ | Scaling factor |
| $K_t$ | Stress concentration factor |
| $L, \lambda, \alpha_l$ | Loss function, regularization coefficient, learning rate |
| $M, \mathcal{M}$ | Extremum and set of all extrema of a given signal |
| $m_i(t), \bar{\gamma}$ | The $i^{th}$ spectral moment of PSD $G(f)$, $i = 0, 1, 2, 4$, and the irregularity factor |
| $N, N_j$ | Size of the full dataset, the training, CV, and test set, $j = tr, CV, test$ |
| $n_i(t)$ | Number of loading cycles of amplitude $S_{a,i}$ counted by time $t$ |
| $N_k$ | Number of amplitudes considered in the rainflow counting method |
| $S_{a,i}$ | Stress amplitude that leads to failure after $N_{f,i}$ loading cycles |
| $x(t), \chi(t), f_{max}$ | Measured and normalized stochastic signal, maximal constructing frequency |
| $x_m, \bar{x}$ | Actual and measured mean loading signal |